\documentclass[letterpaper, 10 pt, conference]{ieeeconf}  

\IEEEoverridecommandlockouts                   

\usepackage{graphics} 
\usepackage[demo]{graphicx}
\usepackage{mathtools}
\usepackage{bm}
\usepackage{epsfig} 
\usepackage{times} 
\usepackage{amsmath} 
\usepackage{amssymb}  
\usepackage{nccmath}
\usepackage{hyperref}
\usepackage{ragged2e}
\usepackage{enumerate}
\usepackage{cancel}
\usepackage{xcolor}
\usepackage{cite}

\newcommand{\vect}[1]{\boldsymbol{#1}}

\allowdisplaybreaks

\title{\LARGE \bf
Design of a Jumping Control Framework with Heuristic Landing \\
for Bipedal Robots
}

\author{Jingwen Zhang$^1$, Junjie Shen$^1$, Yeting Liu$^1$, and Dennis Hong$^1$
\thanks{
$^1$Jingwen Zhang, Junjie Shen, Yeting Liu, and Dennis Hong are with the Robotics and Mechanisms Laboratory, the Department of Mechanical and Aerospace Engineering, University of California, Los Angeles, CA 90095, USA 
{\tt\small \{zhjwzhang, junjieshen, liu1995, dennishong\}@ucla.edu}
}
}

\begin{document}

\maketitle
\thispagestyle{empty}
\pagestyle{empty}

\begin{abstract}

Generating dynamic jumping motions on legged robots remains a challenging control problem as the full flight phase and large landing impact are expected. Compared to quadrupedal robots or other multi-legged robots, bipedal robots place higher requirements for the control strategy given a much smaller footprint. To solve this problem, a novel heuristic landing planner is proposed in this paper. With the momentum feedback during the flight phase, landing locations can be updated to minimize the influence of uncertainties from tracking errors or external disturbances when landing. To the best of our knowledge, this is the first approach to take advantage of the flight phase to reduce the impact of the jump landing which is implemented in the actual robot. By integrating it with a modified kino-dynamics motion planner with centroidal momentum and a low-level controller which explores the whole-body dynamics to hierarchically handle multiple tasks, a complete and versatile jumping control framework is designed in this paper. Extensive results of simulation and hardware jumping experiments on a miniature bipedal robot with proprioceptive actuation are provided to demonstrate that the proposed framework is able to achieve human-like efficient and robust jumping tasks, including directional jump, twisting jump, step jump, and somersaults.

\end{abstract}

\section{Introduction}

With significant progress being made in recent decades, it has been proven that legged robots have the potential to go anywhere humans can go and do whatever humans can do. To fulfill this great potential, dynamic jumping is another required capability besides walking and running. However, apart from higher actuation requirements for the torque density and speed, dynamic jumping control yet remains a challenging problem since it involves a long flight phase where the floating base is uncontrollable without contacts and a large landing impact is expected which requires a more robust control strategy.

Early studies on legged robotic jumping are significantly influenced by Raibert's single-leg hopping machine with a heuristic controller \cite{raibert1984hopping}. Using a similar controller, Hyon and Mita designed another one-legged hopping robot that had an articulated leg composed of three links \cite{hyon2002development}. More recently, model-based methods gain more and more attention. In \cite{de2010feature, wensing2013generation}, the whole-body dynamic model is used in the low-level whole-body control. With commanding the liftoff velocity, simple jumping can be achieved. To generate versatile jumping with a longer horizon, manually finding the trajectory is nearly impossible due to the high degrees of freedom (DoF) of legged robots, especially bipedal robots. Directly using the whole-body model for planning, the robot can produce more intricate behaviors \cite{jump_pro, nguyen2021contact, nguyen2022continuous}. However, due to the complexity of high-dimensional models, these problems sometimes end up being intractable \cite{ddp2, jump_pro, zhang}. In \cite{wensing2013high, wensing2014development}, the spring-loaded inverted pendulum (SLIP) is accepted as the simplified model to plan running and jumping motions. Despite its success, the point mass is considered with the angular momentum being ignored. Jumping motions involving body rotations, which is fairly normal in animals and humans, is hard to accomplish with it. To mitigate this issue, the single rigid body model (SRBM) is a potential solution. \cite{chignoli2021online} successfully implemented it on a quadruped robot for aerial motion trajectory optimization with the assumption of mass-less legs. Unfortunately, bipedal robots require more actuated joints for each leg and non-point feet for active balancing. Basically, the mass-less leg assumption is easily violated for most bipedal robots. Recently, many approaches \cite{dai2014whole, ponton2016convex, ponton2018time, grimminger2020open, chignoli2021humanoid} consider a more versatile kino-dynamic planner for bipedal robots based on the centroidal dynamics \cite{orin2013centroidal} which is an exact projection of the whole-body dynamics. Centroidal dynamics efficiently introduce the angular momentum into the planner, which benefits arbitrary jumping motion generation.

\begin{figure}[!t]
    \centering
    \includegraphics[scale=0.4, trim={0 0 0 -20}]{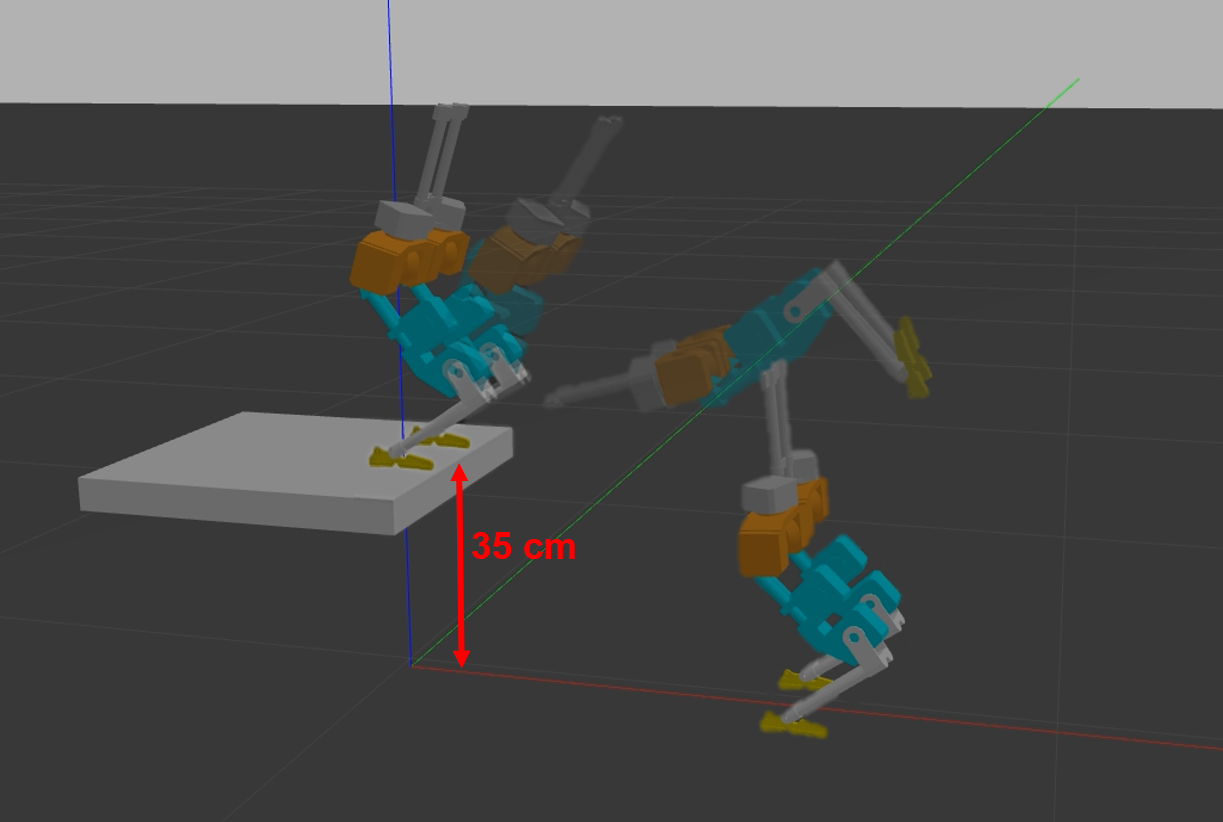}
    \caption{Somersault of BRUCE off of a 35 cm platform.}
    \label{fig:bruce-somersault}
\end{figure}

For landing stabilization, three different methods can be considered: 1) heuristic local damping control \cite{kim2007landing, jo2021robust}, 2) unique structure design of feet to handle impact passively \cite{nabi}, 3) actively consider the whole-body dynamics between the robot and the environment. In the last decade, the solution has converged to the last one with a certain class of optimization-based whole-body control \cite{wensing2022optimization}, which finds optimal solutions to motor commands online based on real-time feedback via solving a convex quadratic program (QP). For quadrupedal robots, this technique is enough to guarantee the safety \cite{nguyen2022continuous, chignoli2021humanoid} if the robot deviates from the optimal trajectory when leaving the ground and is even disturbed in the air since 4 legs provide enough support regions for recovery when landing. For bipedal robots, the tolerance is much lower given a smaller footprint. Besides improving the robustness of the low-level controller, the large landing impact can actually be reduced by adjusting landing locations, which is often ignored in the legged robotic community. The subconscious motion for humans when pushed forward during jumping is moving two legs forward to reduce the influence of the unexpected disturbance on landing. Inspired by this, a heuristic landing planner based on real-time momentum feedback is proposed in this paper. Once the robot leaves the ground, it updates the desired landing locations for feet at high rates using the computed linear and angular momentum as the heuristic feedback.

This paper makes the following contributions:
\begin{enumerate}
\item A novel heuristic landing planner is proposed to improve the landing stability via taking advantage of the momentum feedback in the flight phase to minimize the influence of uncertainties from tracking error or external disturbance on landing. 
\item A complete and versatile jumping framework for bipedal robots is provided with implementation details as shown in Figure \ref{fig:bruce-jump-framework}. Combining the model-based method with centroidal dynamics and the heuristic approach, a more natural jumping behavior can be achieved including squatting before the liftoff, body active rotating to compensate for the unexpected angular momentum, and lowering the body to buffer landing impact.
\item Demonstration of the proposed framework on a miniature bipedal robot that can achieve a variety of jumping tasks, such as directional jump, twisting jump, step jump, and somersaults.
\end{enumerate}

\section{System Overview}
\subsection{BRUCE}
\begin{figure}[!t]
    \centering
    \includegraphics[scale=0.38, trim={0 0 0 -25}]{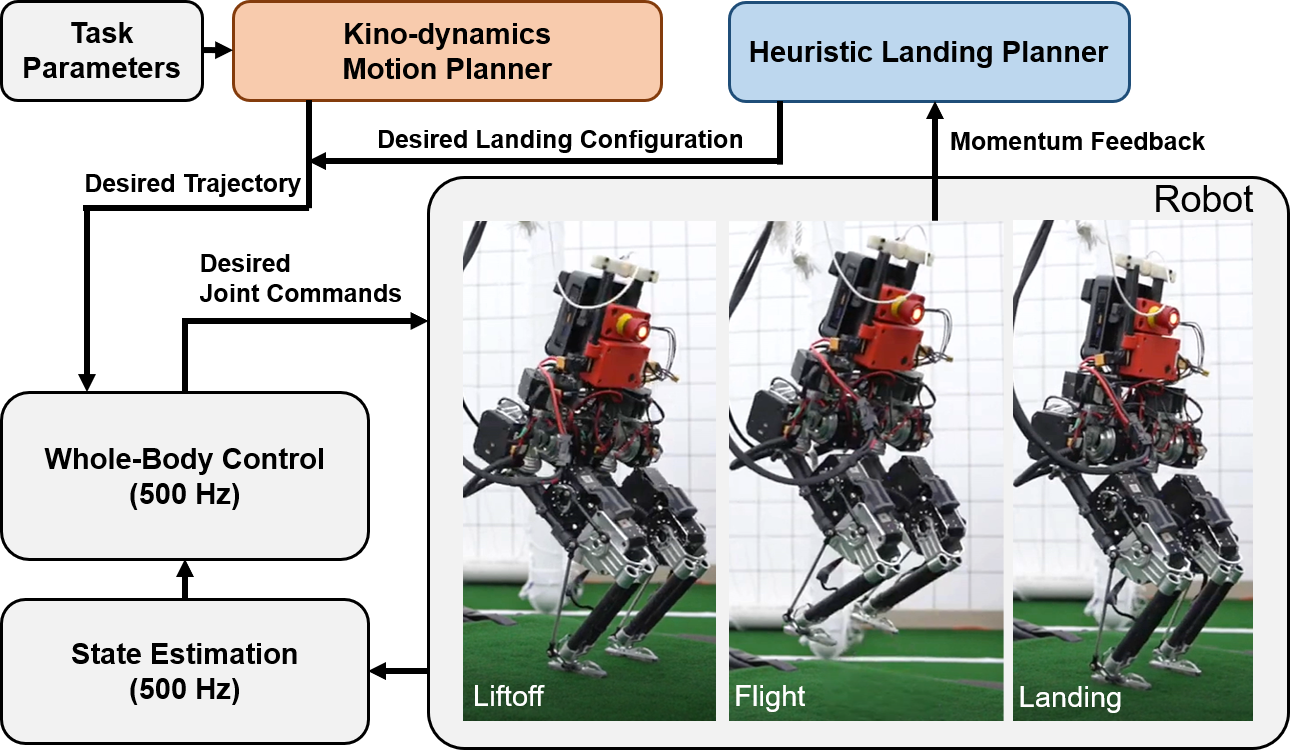}
    \caption{BRUCE jumping framework. Kino-dynamic motion planner generates desired jumping trajectory based on user-selected task parameters including jumping distance/height, twisting angle, contact sequences, etc. Heuristic landing planner updates desired landing locations with real-time momentum feedback in the air. The low-level whole-body controller feeds desired joint commands to the actual robot with high-level commands.}
    \label{fig:bruce-jump-framework}
\end{figure}

To promote bipedal robotic research and improve the accessibility to bipedal robot platforms with dynamic capabilities, the next-generation miniature Bipedal Robot Unit with Compliance Enhanced (BRUCE) has been developed in our previous work \cite{liu2022design} using proprioceptive actuators. For BRUCE, each leg has 5 degrees of freedom (DoF), which includes a spherical hip joint, a knee joint and an ankle joint. To lower the leg inertia, a cable-driven differential pulley system and a linkage mechanism are applied to the hip and ankle joints, respectively. Being a miniature bipedal robot, BRUCE is designed to be approximately 1/3 of an adult male's height, which is around 660 mm. As a result, link lengths of BRUCE and also other major mechanical parameters are summarized in Table \ref{tab:bruce-mech-parameters}.

\begin{table}[!t]
    \centering
    \caption{BRUCE Mechanical Parameters}
    \begin{tabular}{c|c||c|c}
    \hline\hline
        \textbf{Parameter} & \textbf{Value [Unit]} & \textbf{Parameter} & \textbf{Value [Unit]} \\\hline
        Body mass $m_b$ & 1687 [g] & Total mass $m$ & 5118 [g] \\\hline
        Hip mass $m_h$ & 689 [g] & Pelvis length $l_p$ & 150 [mm] \\\hline
        Thigh mass $m_t$ & 889.5 [g] & Thigh length $l_t$ & 175 [mm] \\\hline
        Calf mass $m_c$ & 113 [g] & Calf length $l_c$ & 169.5 [mm] \\\hline
        Foot mass $m_f$ & 24 [g] & Foot length $l_f$ & 24 [mm]\\
    \hline\hline
    \end{tabular}
    \label{tab:bruce-mech-parameters}
\end{table}

To enable BRUCE to detect when the contact between the foot and the ground is created or broken for state estimation purposes, a contact sensor is designed. Tactile switches are embedded into the attached rubber underneath each aluminum foot. To improve the detection performance, one switch is located near the toe with another one near the heel. In this way, contact detection is more robust and sensors are fully protected from the outside environment. Furthermore, all electronics are integrated into the body of BRUCE for fully untethered control. An Intel NUC with Intel Core i5 CPU is used as the onboard PC. The running time is around 20 minutes with a 14.8 V 2200 mAh LiPo battery.

\subsection{Software Architecture}
To make BRUCE favorable to dynamic behaviors which require fast response, the overall software framework is developed in a multithreaded environment, which includes a motor communication thread, a state estimation thread combined with robot model computation, and a feedback control thread as illustrated in Figure \ref{fig:bruce-jump-framework}. The control thread is using whole body controller described in Section \ref{sec:low-level}, which takes the reference trajectories and robot state feedback and computes desired joint torques at a rate of 500 Hz. The reference trajectories come from high-level planners described in Sections \ref{sec:planning} and \ref{sec:heuristic}. Data communication utilizes a custom shared memory library, similar to the setup developed in \cite{nabi}. All programs are implemented in Python while some parts, including kinematics, dynamics, and state estimation, are precompiled using Numba \cite{numbda} for acceleration.

Reliable state estimation is crucial to the good performance of legged systems. In the state estimation thread, a complementarity filter is applied. For body orientation and angular velocity, IMU sensor (ISM330DHCX) readings after proper filters are used. Once the body orientation is determined, IMU accelerometer readings can be integrated to get body velocity and further position. However, these two quantities diverge easily due to sensor noise. Motor encoder readings are introduced as a complement. In specific, the state estimator makes use of the joint encoders for stance leg kinematics to calculate the body position and velocity as a reference. Although Kalman filter is widely used for legged state estimation \cite{est1, est2}, complementarity filter works as well in practice with a much simpler implementation \cite{ding}. Note that this simple approach still comes with some practical issues, e.g., yaw drift, global position inaccuracy.

\section{Motion Planning with Centroidal Momentum} \label{sec:planning}

Bipedal robots have limited supporting regions to recover themselves when landing due to their small footprint. This requires the jumping trajectory must match well with the robot. As a result, the model selection for the planning is crucial. The principle of selection is to make it as simple as possible while keeping versatility to some extent since too complicated models like the full-body dynamics may increase the computational cost significantly and even lead to an intractable problem. In order to achieve a variety of jumping motions, including twisting jumps and somersaults where large body rotations may be required, the point-mass model like the Spring-loaded Linear Inverted Pendulum  (SLIP) model is not considered here.

For comparison, the Single-Rigid-Body Model (SRBM) is used at first. By lumping all the link inertia together to get the fixed local inertia of the represented single rigid body, BRUCE fails to accomplish a stable landing due to the unexpected rotation of the body in the air. It turns out that the unexpected angular momentum comes from the thigh rotation when trying to lift off from the ground but it is ignored inside the SRBM. To tackle this issue, we accept centroidal dynamics \cite{orin2013centroidal} for motion planning. Since centroidal dynamics consider the full-body mass and inertia distribution, the optimized jumping motion can compensate for the unexpected angular momentum from leg movements. 

\subsection{Decision Variables}
For the full-body mass and inertia distribution, the joint configuration must be considered. As a result, after including the extra 6 DoFs from the floating body in the joint states, the decision variables are determined for the planner with centroidal momentum as 

\begin{align}
    \vect{\Gamma} = \{ & \vect{q}[k], \vect{v}[k], \vect{r}[k], \vect{\dot{r}}[k], \vect{\ddot{r}}[k], \vect{F_j}[k], \vect{h}[k], \nonumber \\
    & \text{for all time instances $k$} \}
\end{align}

where $\vect{q}$ and $\vect{v}$ denote joint positions and velocities including the floating base. For BRUCE with 10 active joints, $\bm{q} \in \mathcal{R}^{7+10}$, and $\bm{v} \in \mathcal{R}^{6+10}$ where $\bm{q}$ included 7 variables for the floating-base since quaternion is chosen to represent the orientation which requires 4 instead of 3 to avoid the gimbal locking issue. $\vect{r}, \vect{\dot{r}}, \vect{\ddot{r}}$ represent CoM states including positions, velocities, and accelerations. $\vect{F_j}$ is the contact forces for the $j^{th}$ contact point. Note that only point contact is considered here. To avoid losing generality, any type of contact can be represented with the point contact. For example, the line contact or square face contact can be divided into 2 or 4-point contacts on the edge. $\vect{h}$ describes the angular components in the centroidal momentum as defined in \cite{orin2013centroidal} which is the exact projection of all the link momentum on the CoM coordinate. 

Other works \cite{dai2014whole, ponton2018time} might consider additional decision variables, such as the contact location, the unscheduled contact sequence, and the time step $dt$. Here, they are fixed for simplification.

\subsection{Constraints}

As a kino-dynamics planner, both dynamics and kinematics are considered. The motion of equations for the model with centroidal momentum is written as:
\begin{align}
    m\vect{\ddot{r}}[k] = \sum^{}_{j} \vect{F_j}[k] + m\vect{g} &  \label{const:linear mom. rate} \\
    \vect{\dot{h}}[k] = \sum^{}_{j} (\vect{c_j}[k] - \vect{r_j}[k]) \times \vect{F_j}[k] & \label{const:angular mom. rate} \\
    \vect{h}[k] = \vect{A}(\vect{q}[k])\vect{v}[k] & \label{const:angular mom.}
\end{align}
where $\bm{c_j}$ denotes the pre-specified contact location for the $j^{th}$ contact point. As shown in Equation \eqref{const:angular mom.}, the centroidal angular momentum is connected with the joint states with the Centroidal Momentum Matrix (CMM) as defined in \cite{orin2013centroidal}. Although being nonlinear, the computation of CMM can be achieved efficiently \cite{wensing2016improved}. Due to the introduction of joint states, the full-body kinematic must be added to ensure kinematic consistency.
\begin{align}
    \vect{r}[k] = f_{com}(\vect{q}[k]) & \label{const:CoM FK pos} \\
    \vect{c}[k] = f_{contact}(\vect{q}[k]) &  \label{const:contact FK pos}
\end{align}
where the function $f(\cdot)$ represents the corresponding forward kinematics for the CoM positions and contact locations. To ensure the consistency of the generated trajectory, integration constraints are formulated as follows:
\begin{align}
    \vect{q}[k] - \vect{q}[k-1] = \vect{v}[k]dt &  \label{const:joint integration}\\
\vect{h}[k] - \vect{h}[k-1] = \vect{\dot{h}}[k]dt &  \label{const:CoM integration1}\\
\vect{r}[k] - \vect{r}[k-1] = \vect{\dot{r}}[k]dt &  \label{const:CoM integration2}\\
\vect{\dot{r}}[k] - \vect{\dot{r}}[k-1] = \vect{\ddot{r}}[k]dt & \label{const:CoM integration3}
\end{align}

When the contact is active for the $j^{th}$ contact point, the contact force is limited inside the friction cone to avoid sliding,
\begin{equation}
    \sqrt{(\vect{F_j})^2_x + (\vect{F_j})^2_y} \leq \mu(\vect{F_j})^2_z, (\vect{F_j})_z \geq 0 \label{const:friction cone}
\end{equation}

Until here, all constraints describe the general motion of the model with centroidal momentum. To meet the task-specific requirements, decision variables are constrained in a pre-defined boundary set $\mathcal{Q}$ as boundary conditions.
\begin{equation}
    \vect{\Gamma} \in \mathcal{Q} \label{const:boundary conditions}
\end{equation}

In the jumping optimization, besides physical boundaries on numerical values of decision variables, users can define the desired jumping parameters in the boundary condition, for example,
\begin{align}
   & \vect{r}[1] = \vect{r}_0, \vect{r}[N] = \vect{r}_0  \label{const:jump1} \\
     & (\vect{r}[k])_z \leq h_{nom} \quad \text{when in stance} \label{const:jump2} \\
     & h_{nom} \leq (\vect{r}[k])_z \leq h_{max} \quad \text{when in flight} \label{const:jump3}\\
     & \vect{\dot{r}}[k_i] = \vect{v}_{lo} \quad \text{$k_i$ is the time the robot lifts off}  \label{const:jump4}
\end{align}
where $\bm{r}_0$ denotes the COM positions when the robot is standing still, $h_{nom}$ is the nominal height when the robot is on the ground, $h_{max}$ is the maximum COM height, and $\bm{v}_{lo}$ denotes the liftoff velocity so that users can change its value to reach different jumping height. Similarly, if a twisting jump is desired, constraints on the body orientation and angular momentum can be added to limit the twisting angle and velocity.

\subsection{Complete Formulation}
The complete problem can be formulated as a nonlinear programming (NLP) as follows:
\begin{subequations}\label{mp:bruce-jump-cm-NLP}
\begin{align}
    \underset{\Gamma}{\textbf{min}} \quad \sum^{N}_{k=1}
    &
    \bigg( \left\|\vect{q}[k] - \vect{q_{nom}}[k] \right\|^2_{\vect{Q_q}} + \left\|\vect{v}[k]\right\|^2_{\vect{Q_v}} + \left\|\vect{\ddot{r}}[k] \right\|^2 \nonumber \\ 
    &+ \left\|\vect{\dot{h}}[k]\right\|^2_{\vect{Q_h}} + \sum_{j}\left\|\vect{F_j}[k]\right\|^2_{\vect{Q_f}} \bigg) dt \tag{\ref{mp:bruce-jump-cm-NLP}}
\end{align}
\begin{align}
\text{s.t., for each knot point $k = 1, \ldots, N$ and} & \nonumber\\
\text{for each contact point $j = 1, \ldots, M$} & \nonumber \\
\eqref{const:linear mom. rate}, \eqref{const:angular mom. rate}, \eqref{const:angular mom.}, \eqref{const:CoM FK pos}, \eqref{const:contact FK pos}, \eqref{const:joint integration}, \eqref{const:CoM integration1}, \eqref{const:CoM integration2}, \eqref{const:CoM integration3}, \eqref{const:friction cone}, \eqref{const:boundary conditions} & &
\end{align}
\end{subequations}

In the cost function, $\left\| \cdot \right\|^2_{\bm{Q}}$ is the abbreviation for the quadratic cost with the weight matrix as $\bm{Q} \geq 0$. The terminal cost is ignored. Instead, the final state is put as an additional hard constraint in Constraint (\ref{const:boundary conditions}). And new running cost terms on joint positions and velocities are introduced. The difference between an initial guess $\bm{q_{nom}}$ and optimized joint positions is considered since it does not make any physical meaning to penalize purely large joint positions. $\bm{q_{nom}}$ can be defined as a fixed nominal position which can be the safest configuration for the robot or a whole trajectory along the time span from users' educated guesses or other simple planners.

\section{Heuristic Landing} \label{sec:heuristic}
Besides the unexpected angular momentum generated from leg movements, another source of uncertainty comes from the tracking error. The low-level controller may not be able to track the optimized jumping trajectory perfectly in practice. Although in \cite{nguyen2022continuous} and \cite{chignoli2021humanoid}, model predictive control (MPC) is combined with the whole-body controller to improve the tracking performance and disturbance rejection, the stable margin is still small for bipedal robots. To tackle this problem, a landing planner is proposed here. The intuition behind this is that the robot is capable of adjusting its foot freely once the robot leaves the ground. If a human tries to jump in-situ but is pushed forward in the air, the subconscious reaction is to move both legs forward to catch the landing impact. Similarly, the robot can update the landing locations even if the state of the robot is not exactly the same as planned when leaving the ground or being disturbed in the air.

The ideal approach to update the landing locations may be model-based optimization techniques. However, the flight phase may be often too short to apply models considering momentums like the SRBM and centroidal dynamics. Finding the optimal solution to complex models in a real-time manner still remains an open question. Meanwhile, the point-mass models are too abstract to capture the requirements of computing the optimal landing locations. Inspired by the Raibert heuristic for hopping \cite{raibert1984hopping} and the capture point \cite{pratt2006capture}, the momentum of the robot in the air is used as the heuristic to update the landing locations. For example, if the robot is leaving the ground with a non-zero angular momentum along the y direction, the foot must be moved forward/backward in the air. The landing locations can be updated based on momentum feedback as follows:
\begin{align}
    & p^x = p^x_{nom} + W^x_l l^x + W^x_k k^y \\
    & p^y = p^y_{nom} + W^y_l l^y + W^y_k k^x 
\end{align}
where $p^x$ and $p^y$ denote the updated foot x and y positions while the z position still follows the optimized trajectory in the air, $p^x_{nom}$ and $p^y_{nom}$ are the nominal positions which can be set by users or obtained from the optimized trajectory, $l^x$ and $l^y$ are the linear momentum feedback along x and y directions while $k^x$ and $k^y$ denote the angular momentum feedback, and $W$ is the heuristic gain. 

Additionally, inspired by the human behavior to increase the y clearance between two legs when being pushed sideways so as to have a larger supporting region for recovery when landing, an additional term can be added to $p^y$ to adjust the y distance between two legs.
\begin{equation} \label{eqn:landing-y-clearance}
    \Delta y = W_c |l^y|
\end{equation}
Lastly, it is easy for the direct heuristic planner to find a landing location out of the leg's reachable region. To avoid this, the heuristic landing planner is designed with a saturation function as follows:
\begin{equation} \label{eqn:land-planner-x}
    p^x_{des} = \begin{cases}
    p^x & \text{if $|p^x| \leq p^x_{max}$} \\
    p^x_{max} \text{sgn}(p^x) & \text{if $|p^x| > p^x_{max}$}
    \end{cases}
\end{equation}
\begin{equation} \label{eqn:land-planner-y}
    p^y_{des} = \begin{cases}
    p^y + (-1)^i \Delta y & \text{if $|\hat{p}^y| \leq p^y_{max}$} \\
    p^y_{max} \text{sgn}(p^y + (-1)^i \Delta y) & \text{if $|\hat{p}^y| > p^y_{max}$}
    \end{cases}
\end{equation}
where $\hat{p}^y = p^y + (-1)^i \Delta y$, $p^x_{max}$ and $p^y_{max}$ denote the maximum x and y position that the leg can reach, and $i = 0$ for the left leg while $i=1$ for the right leg. With this landing planner, the robot can handle both the tracking error when lifting off and the disturbance in the air. More stable landing and push recovery in the air can be found in Section \ref{sec:results}.

\section{Jumping Controller} \label{sec:low-level}
For the low-level control, the goal is to improve the tracking performance along the optimized jumping trajectory. Although the QP-based whole-body controller is able to provide compliant behaviors and strong robustness, it heavily depends on the high quality of the dynamic model which is often difficult to obtain in practice. Additionally, jumping is a highly dynamic motion that requires significant acceleration and control of the fast leg movement is typically hard. However, inverse kinematics approaches only require the robot kinematic model which is much easier to make accurate. Joint-level PD control can benefit humanoid control due to its modeling error compensation and high updating frequency \cite{feng, kim2019highly}. Based on that, a combined low-level jump controller for each joint is utilized to improve the tracking performance as follows:
\begin{align} \label{eqn:joint-controller}
    \tau^{des}=\tau_{ff}+K_p\big(q_{d}-q\big)+K_d\big(\dot{q}_{d}-\dot{q}\big),
\end{align}
where $K_p/K_d$ is the P$/$D feedback gain for each joint. In Equation (\ref{eqn:joint-controller}), the last two terms serve as the joint-level feedback terms where $q_{d}$ and $\dot{q}_{d}$ are desired joint position and velocity that can be obtained by solving the leg inverse kinematics. The first term is treated as the feedforward term from the operational-space controller. In jumping control, the feedforward term is computed separately for different phases. When the robot is in the air without contact, the robot is following the ballistic trajectory under gravity. The PD feedback terms in Equation \eqref{eqn:joint-controller} are enough to control the foot position as commanded by the heuristic landing planner in Section \ref{sec:heuristic}. $\tau_{ff}$ is set as 0 in the air accordingly.

For the liftoff and landing phase where the ground contact is active for the robot, a weighted hierarchical whole-body controller is formulated as a quadratic programming (QP) as follows:
\begin{subequations}\label{wbc_qp}
\begin{align}
    \min_{\bm{\ddot{q}},\bm{f}_j}~
    &
    \sum_{i=1}^{N_t}\left\|\bm{J}_i\bm{\ddot{q}}+\bm{\dot{J}}_i\bm{\dot{q}}-\bm{\ddot{x}}_i^{des}\right\|_{\bm{W}_i}^2\nonumber\\
    &+\sum_{j=1}^{N_c}\left\|\bm{f}_j\right\|_{\bm{W}_f}^2+\left\|\bm{\ddot{q}}\right\|_{\bm{W}_{\ddot{q}}}^2\label{wbc_cost} \tag{\ref{wbc_qp}} \\
    \textrm{s.t.}~
    &~\bm{H}_b\bm{\ddot{q}}+\bm{C}_b\bm{\dot{q}}+\bm{G}_b-\sum_{j=1}^{N_c}\bm{J}_{c_j, b}^\top\bm{f}_j=\bm{0},\label{base_dyn}\\
    &~\bm{f}_j\in\bm{\mathcal{C}}_j,~j=1,\cdots,N_c,\label{cone}
\end{align}
\end{subequations}
where $\bm{J}_i$ is the $i$th task Jacobian and $N_t$ is the number of tasks.
As we can see, the $i$th operational task is set as a QP cost with priority implicitly being enforced with weight $\bm{W}_i$. In addition to the task costs, regularization costs are added to the decision variables $\bm{\ddot{q}}$ and $\bm{f}_j$ with small weights $\bm{W}_{\ddot{q}}$ and $\bm{W}_f$ respectively to ensure the overall QP cost is strictly positive definite even when the task Jacobians contain singularities, which avoids potential numerical issues. In Problem (\ref{wbc_qp}), only the floating-base components of the full-body dynamics are used. After solving the optimal accelerations and forces, the joint torques can be retrieved using the joint components of the dynamics as follows: 
\begin{align}\label{opt_torque}
    \bm{\tau}_{ff}=\bm{H}_j\bm{\ddot{q}}^*+\bm{C}_j\bm{\dot{q}}+\bm{G}_j-\sum_{j=1}^{N_c}\bm{J}_{c_j, j}^\top\bm{f}_j^*
\end{align} 
In this manner, variables for $\bm{\tau}$ can be removed from the decision variables to accelerate the QP solving. But we always assume enough torque that the actuator can provide, i.e., no torque limits. In order to track the optimized jumping trajectory, multiple tasks are prioritized in the following sequence, e.g., the $1^{st}$ task means the highest priority with the largest task weight $\bm{W}_1$).

\subsection{Task 1 - Stance Leg}
In general, to ensure the stance leg is nonmoving, besides Constraint \eqref{cone} ensures each contact force is bounded and lies within the local friction cone $\bm{\mathcal{C}}_j$ which is approximated by a square pyramid for linearity, the contact acceleration is also fixed to zero as a hard constraint with the equation $\bm{J}_{c_j}\bm{\ddot{q}}+\bm{\dot{J}}_{c_j}\bm{\dot{q}}=\bm{0}$. However, it was treated as the first task, i.e., a soft constraint with sufficiently large task weight and $\ddot{\bm{x}}^{des}_1 = \bm{0}$. This can speed up the QP and give better numerical stability \cite{feng}.

\subsection{Task 2 - Linear Momentum}
In particular, the linear momentum task consists of both feedforward and feedback terms, which are specified in the form of
\begin{equation}
    \bm{\ddot{x}}_2^{des}=\bm{a}_2^{ref}\!+\!\bm{K}_p\big(\bm{p}_2^{ref}-\bm{p}_2\big)\!+\!\bm{K}_d\big(\bm{v}_2^{ref}-\bm{v}_2\big) \label{linear}
\end{equation}
where $\bm{a}_2^{ref}$, $\bm{v}_2^{ref}$, $\bm{p}_2^{ref}$ are the linear acceleration, velocity, and position from the optimized jumping trajectory in Section \ref{sec:planning}, and $\bm{K}_p/\bm{K}_d$ is the proportional$/$derivative (P$/$D) feedback gain matrix. The linear components of the centroidal momentum matrix (CMM) are used as the task Jacobian.

\subsection{Task 3 - Torso Orientation}
Controlling the torso orientation is essential for the angular momentum compensation during jumping. The task acceleration for all three angles are described as:
\begin{equation}
    \bm{\ddot{x}}_3^{des}=\bm{\alpha}_3^{ref}\!+\!\bm{K}_p\textrm{Log}\big(\bm{R}_3^\top\bm{R}_3^{ref}\big)\!+\!\bm{K}_d\big(\bm{\omega}_3^{ref}-\bm{\omega}_3\big)
\end{equation}
where $\bm{\alpha}_3^{ref}$, $\bm{\omega}_3^{ref}$, $\bm{R}_3^{ref}$ are the desired angular acceleration, velocity, orientation, and the logarithm operator $\textrm{Log}:\textsf{SO}(3)\to\mathbb{R}^3$ converts a rotation matrix to its corresponding axis–angle representation. In practice, $\bm{\alpha}_3^{ref}$ is set to $\bm{0}$ since it is hard to define angular acceleration while $\bm{\omega}_3^{ref}$ and $\bm{R}_3^{ref}$ can be obtained from the optimized trajectory.

\subsection{Task 4 - Angular Momentum}
With active orientation control in Task 3, the angular momentum task is of low priority yet to regularize the rotation of the body. As a result, the angular momentum task is to damp out excessive angular momentum:
\begin{equation}
    \bm{\ddot{x}}_4^{des}=-\bm{K}_d\bm{k}
\end{equation}
The angular components of the CMM are used as the task Jacobian.

\section{Results} \label{sec:results}
In this section, various jumping tasks of simulation and hardware experiments for BRUCE are conducted to verify the capability of the proposed dynamic jumping framework. The open-source simulator Gazebo \cite{koenig2004design} with the ODE physics engine is used as the simulation environment. All the trajectories of the following jumping tasks are generated using the jumping planner with centroidal momentum in Section \ref{sec:planning}. With warm start techniques, the solving time varies from around 10 sec to 5 mins depending on the complexity of the task using the SNOPT solver \cite{gill2005snopt} in Drake \cite{drake2019}. To update landing locations in the air, the foot trajectory along x and y directions are updated with the interpolation of the desired foot locations from the landing planner in Section \ref{sec:heuristic} while it is still following the optimized trajectory along the z direction. The low-level jumping controller tracks the modified trajectory with the off-the-shelf QP solver OSQP \cite{osqp} which can achieve a 500 Hz updating frequency, sufficient for real-time feedback control. All of the following experiments can be seen in the accompanied video. 

\subsection{Basic Jumping}
\begin{figure}[!t]
    \centering
    \includegraphics[scale=0.4, trim={-10 80 0 60}]{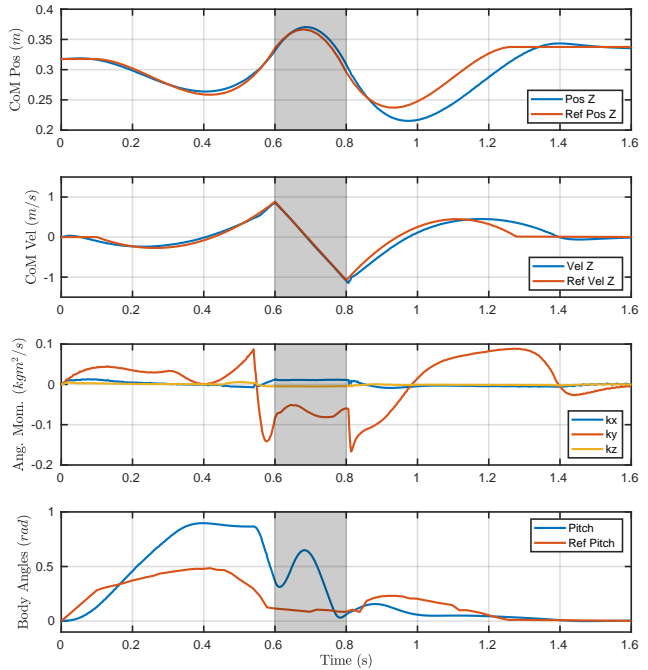}
    \caption{In-situ jumping trajectory for BRUCE. Shaded regions represent the flight phase.}
    \label{fig:bruce-insitu-jump-curve}
\end{figure}

To verify the capability of the jumping motion planner, the trajectories for different basic jumping tasks including in-situ jump, directional jump, and twisting jump are generated offline. For these tasks, only minor changes are required for the Constraint \eqref{const:boundary conditions} in Problem \eqref{mp:bruce-jump-cm-NLP}, e.g., the twisting angle, desired COM final positions, etc. As shown in Figures \ref{fig:bruce-jump-framework} and \ref{fig:bruce-insitu-jump-curve}, BRUCE is able to accomplish a natural jumping and land stably. The active rotation of BRUCE's body during the liftoff phase can be noticed. Since the centroidal dynamics consider leg inertia which is dependent on joint configuration, the planner optimizes the body rotation to compensate for the angular momentum generated from leg movements. Actually, this is exactly what human does when jumping. In order to jump higher, humans would lean their body forwards when squatting and then suddenly rotate the body backward to lift off. With the proposed framework, BRUCE is able to achieve a more natural in-situ jumping motion which involves squatting before the liftoff and lowering the body to relieve the impact after landing. 

Meanwhile, the QP-based whole-body controller \eqref{wbc_qp} is able to track the COM position and velocity very well in practice with a large linear momentum task weight as shown in Figure \ref{fig:bruce-insitu-jump-curve}. However, due to the small task weight on the torso orientation tracking, even with the optimized body rotation for angular momentum compensation, the angular momentum along the y direction is still not negligible leading to decreasing pitch angle of the body in the air. With the proposed landing planner in the air, the robot is still able to land safely with adjusted landing locations. Note that the body pitch angle is first increasing although with a decreasing trend. This is because the landing planner commands both legs to move forwards. In the ideal case, the body orientation can be kept still. However, the inertia of the body is not big enough compared to the leg inertia for BRUCE. As a result, the body will rotate instead of purely moving legs due to the conservation of angular momentum.

The directional jump and the twisting jump are also conducted in both the simulation environment and the actual robot hardware as shown in the accompanied video. To report here, the maximum jumping height (COM z position change) is around 15 cm (22.7\% of its total height). The maximum jumping distance and twisting angle are around 10 cm and 30 deg. The main limitation here is that the desired landing locations cannot be reached with respect to the global frame due to the accompanied body rotation described previously.

\subsection{Step Jumping}
\begin{figure}[!t]
    \centering
    \includegraphics[scale=0.3, trim={35 0 100 0}, angle=270]{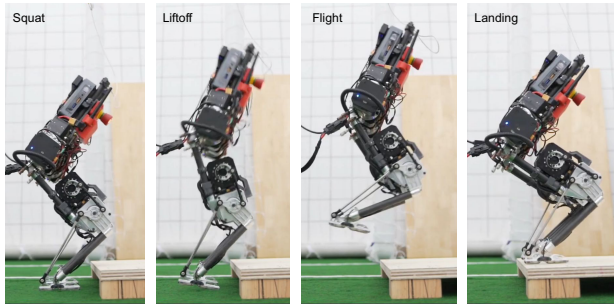}
    \caption{Screeshots of BRUCE step jumping onto a 5cm step. Motion sequence starts from squatting, then lifting off, to the flight phase, ends with a stable landing.}
    \label{fig:bruce-step-jump}
\end{figure}

By combining the basic jumping motions, BRUCE is also capable of jumping onto a 5cm step and jumping downwards as shown in the accompanied video with the proposed jumping framework. Note that the step height is chosen to be conservative in this experiment. In Figure \ref{fig:bruce-step-jump}, BRUCE leans the body forward when squatting and then extends the leg while rotating backward to lift off. As a result, a large portion of the angular momentum generated from the leg movement is compensated while the heuristic landing planner adjusts the landing location in the flight phase to balance out the remaining nonzero angular momentum. When landing, the robot tends to lower the body first to buffer the impact. And eventually, it recovers to the nominal configuration and prepares for the next jump on the step.

\subsection{Push Recovery}
To verify the robustness of the landing planner proposed in Section \ref{sec:heuristic}, a push recovery test is conducted. In the simulation, BRUCE is commanded to jump in-situ and a constant 70 N pushing force with a duration of 0.01 s is respectively applied to the torso of the robot along x and y directions after 0.03 s in the air. Note that the flight phase is only around 0.2 s. Due to the choice of momentum as the heuristic, the landing planner is able to compute the desired landing positions very fast and leave enough time for the low-level controller to move its legs.

\begin{figure}[!t]
    \centering
    \includegraphics[scale=0.4, trim={30 50 0 0}]{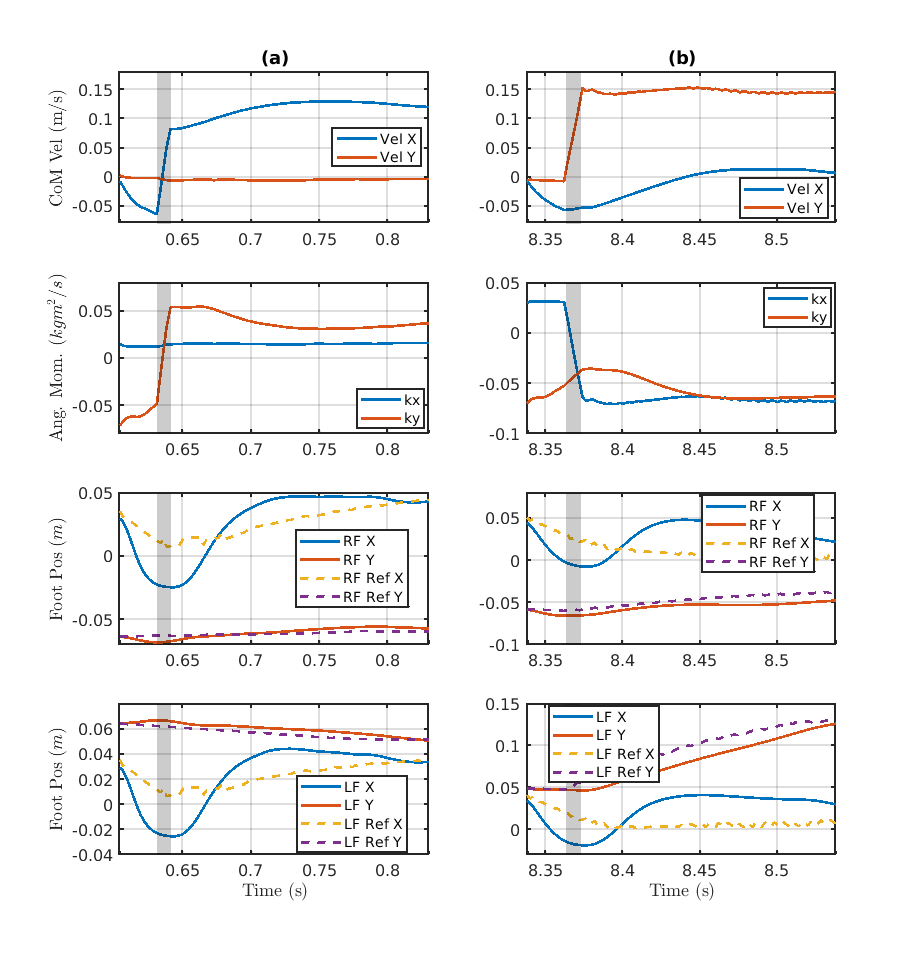}
    \caption{Simulation results of push recovery. The shaded region represents the 0.01s duration of the push while the graph only shows the trajectory of the flight phase. (a) Push along +x direction. (b) Push along +y direction.}
    \label{fig:bruce-push-recovery}
\end{figure}

The simulation results are shown in Figure \ref{fig:bruce-push-recovery}. When the robot is pushed along the +x direction, COM x velocity changes immediately and the angular momentum along the y direction increases since the force is applied on the torso leading to a body rotation. As a result, Equation \eqref{eqn:land-planner-x} commands a forwarding landing position. Similarly, in Figure \ref{fig:bruce-push-recovery}(b),  COM y velocity and the angular momentum along x direction change accordingly when pushed. Since the y-clearance term in Equation \eqref{eqn:landing-y-clearance} is applied to the two legs with a different sign, Equation \eqref{eqn:land-planner-y} moves the left foot along +y direction significantly while the right foot is not changed too much. As a result, the distance between the two legs is becoming larger to increase the capability of the robot to stay balanced when landing along the y direction. However, in both cases, although the foot is already in the commanded position with respect to the body frame, the tracking of the foot position with respect to the global frame is not perfect since the rotation of BRUCE's body is inevitable when commanding to move legs in the air due to the comparable inertia of the body and legs.

\subsection{Somersault}
\begin{figure}[!t]
    \centering
    \includegraphics[scale=0.26]{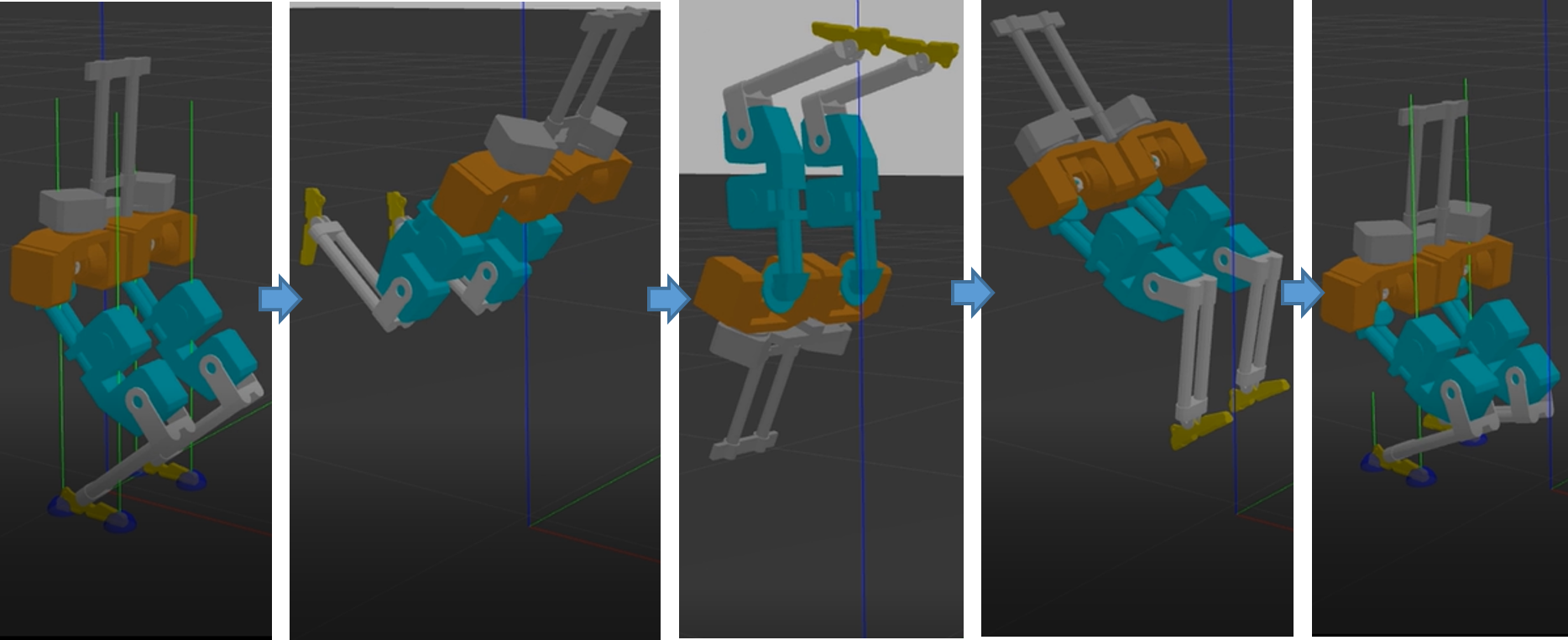}
    \caption{Somersault of BRUCE in-situ with relaxed torque limits.}
    \label{fig:bruce-somersault-insitu}
\end{figure}
To further explore the dynamic capability of BRUCE and the potential of the proposed jumping framework, a somersault is achieved in the simulation using the proposed approach. Due to the quaternion representation of the orientation in Problem \eqref{mp:bruce-jump-cm-NLP}, the planner is able to deal with the task that requires the body of BRUCE to rotate 360 deg. In Figure \ref{fig:bruce-somersault}, BRUCE executes a somersault to jump off of a 35 cm platform. The actual actuator of BRUCE is not powerful enough to support the somersault in-situ. After relaxing the torque limits in the simulation, BRUCE is able to perform a somersault in-situ as shown in Figure \ref{fig:bruce-somersault-insitu}.

\section{Conclusions and Future Works}
In this paper, a complete dynamic jumping framework for bipedal robots with a novel heuristic landing planner is presented. Specifically, the high-level jumping planner with centroidal momentum is solving a NLP offline to get the local-optimal jumping trajectory, which is a series of motions similar to human jumping including squatting before liftoff, body lowering after landing, and body rotating to compensate for the angular momentum. To deal with the tracking error when lifting off and possible disturbances in the air for safer landing, the heuristic landing planner updates the landing positions in a real-time manner during the flight phase based on the momentum feedback. To the best of our knowledge, this is the first approach to take advantage of the flight phase to reduce the impact of the jump landing which is implemented in the actual robot. The low-level whole-body controller is finding the required joint torques to best accomplish the operational-space tasks by solving a small-scale QP, which guarantees the global optimality and ensures a 500 Hz updating frequency. With this framework, a miniature bipedal robot, BRUCE is capable of directional jumps, twisting jumps, jumps with push in the air, and somersaults, which demonstrates the versatility and robustness of the framework.

In the future, a better performance in the actual hardware can expected if the robot, BRUCE can be upgraded with a more optimized mass distribution. The more inertia lumped into the body, the more benefits the proposed framework can get to control the actual hardware. Due to the long solving time of the jumping planner, an offline motion library like \cite{bjelonic2022offline} is under exploration as well for online dynamic behavior execution.

\section*{Acknowledgment}
This work was partially supported by the Office of Naval Research through grant N00014-15-1-2064.


{
\bibliographystyle{ieeetr}
\bibliography{references}
}

\end{document}